\documentclass[10pt,twocolumn,letterpaper]{article}

\usepackage[pagenumbers]{cvpr} % To force page numbers, e.g. for an arXiv version

\definecolor{cvprblue}{rgb}{0.21,0.49,0.74}
\usepackage[pagebackref,breaklinks,colorlinks,allcolors=cvprblue]{hyperref}

\usepackage{multirow}
\usepackage{bbm}

\def\paperID{9759} % *** Enter the Paper ID here
\def\confName{CVPR}
\def\confYear{2025}

\usepackage{amsmath,amsfonts,bm}

\def\eqref#1{equation~\ref{#1}}
\def\1{\bm{1}}

\def\ri{{\textnormal{i}}}
\def\rj{{\textnormal{j}}}
\def\rk{{\textnormal{k}}}

\def\vb{{\bm{b}}}
\def\vc{{\bm{c}}}
\def\vd{{\bm{d}}}
\def\ve{{\bm{e}}}

\def\vh{{\bm{h}}}

\def\vk{{\bm{k}}}

\def\vo{{\bm{o}}}
\def\vp{{\bm{p}}}
\def\vq{{\bm{q}}}
\def\vr{{\bm{r}}}

\def\vt{{\bm{t}}}

\def\vv{{\bm{v}}}

\def\vx{{\bm{x}}}

\def\vz{{\bm{z}}}

\def\mA{{\bm{A}}}

\def\mI{{\bm{I}}}

\def\mK{{\bm{K}}}

\def\mP{{\bm{P}}}

\def\mR{{\bm{R}}}

\def\mT{{\bm{T}}}

\def\mW{{\bm{W}}}

\DeclareMathAlphabet{\mathsfit}{\encodingdefault}{\sfdefault}{m}{sl}
\SetMathAlphabet{\mathsfit}{bold}{\encodingdefault}{\sfdefault}{bx}{n}

\newcommand{\R}{\mathbb{R}}

\DeclareMathOperator{\Tr}{Tr}

\title{Geometric Point Attention Transformer for 3D Shape Reassembly}

\author{
    Jiahan Li$^*$\\
    Tsinghua University\\
    {\tt\small ced3ljhypc@gmail.com}
    \and
    Chaoran Cheng$^*$\\
    University of Illinois Urbana-Champaign\\
    {\tt\small chaoran7@illinois.edu}
    \and
    Jianzhu Ma\\
    Tsinghua University\\
    {\tt\small majianzhu@tsinghua.edu.cn}\\
    \and
    Ge Liu\\
    University of Illinois Urbana-Champaign\\
    {\tt\small geliu@illinois.edu}
}

\begin{document}
\maketitle

\begin{abstract}
Shape assembly, which aims to reassemble separate parts into a complete object, has gained significant interest in recent years. Existing methods primarily rely on networks to predict the poses of individual parts, but often fail to effectively capture the geometric interactions between the parts and their poses. In this paper, we present the Geometric Point Attention Transformer (GPAT), a network specifically designed to address the challenges of reasoning about geometric relationships. In the geometric point attention module, we integrate both global shape information and local pairwise geometric features, along with poses represented as rotation and translation vectors for each part. To enable iterative updates and dynamic reasoning, we introduce a geometric recycling scheme, where each prediction is fed into the next iteration for refinement. We evaluate our model on both the semantic and geometric assembly tasks, showing that it outperforms previous methods in absolute pose estimation, achieving accurate pose predictions and high alignment accuracy.
\end{abstract}
\section{Introduction}

Shape assembly aims to combine parts or fragments to create a complete 3D object, with applications in fields such as robotics \cite{zakka2020form2fit,zeng2021transporter}, bone reconstruction \cite{yin2011automatic,nasiri2022multiple}, archaeology \cite{cohen2013virtual,huang2006reassembling}, and manufacturing \cite{tian2022assemble}. This task is challenging and requires expertise in understanding mechanical structures and accurately matching components, making it prone to errors. Broadly speaking, shape assembly can be divided into two tasks: semantic part assembly and geometric fractured assembly. Semantic assembly \cite{li2020learning,zhan2020generative,li2024category} involves assembling meaningful parts, such as the legs and handles of a chair, into a complete structure by using both geometric clues to understand part functions and semantic information to determine their positions. In contrast, geometric assembly \cite{sellan2022breaking,wu2023leveraging} focuses on reassembling objects that have been broken by external forces, such as putting together fragments of a bowl, where only geometric information (e.g., shapes and textures) is available, without semantic labels for individual parts.

With the release of large-scale 3D datasets like PartNet \cite{mo2019partnet} and Breaking Bad \cite{sellan2022breaking}, shape assembly methods have evolved from traditional geometric matching based on hand-crafted features \cite{papaioannou2003automatic,huang2006reassembling} to more advanced deep learning approaches \cite{toler2010multi,funkhouser2011learning,li2020learning,zhan2020generative,chen2022neural,narayan2022rgl}. In these modern methods, each part is represented as a point cloud in 3D space. Typically, an encoder extracts part-level features from each point cloud, which are then used to predict the 6-DoF (Degrees of Freedom) pose for each part, including rotation and translation vectors to align the part in its target position. Architectures for predicting these part-level poses include multi-layer perceptrons, LSTMs, and graph neural networks (GNNs) \cite{schor2019componet,sung2017complementme,wu2020pq}. To improve regression accuracy and placement precision, recent approaches also explore techniques such as equivariant representations \cite{wu2020pq}, generative modeling \cite{cheng2023score,scarpellini2024diffassemble,lu2024jigsaw++}, and the application of prior knowledge and post-processing matching methods \cite{li2024category,lu2024jigsaw,lee20243d}.

Despite these advancements, current shape assembly methods still face significant challenges, particularly those that use regression networks to predict the absolute poses of each part based on extracted features \cite{zhan2020generative,wu2023leveraging}. First, in addition to global features from the encoder, it is crucial to accurately model the local geometric relationships between different parts. For instance, when assembling a chair, the legs must be positioned at precise distances and angles relative to the seat; even minor deviations can lead to instability or misalignment. Second, since shape assembly is framed as a part-level 6-DoF prediction problem, the network must explicitly consider both rotation and translation transformations for each part. However, common architectures like multi-layer perceptrons (MLPs) or graph neural networks (GNNs) often fail to effectively capture these 6-DoF features, leading to inaccurate positioning \cite{lee20243d}. For example, accurately predicting the orientation of fragmented bowl pieces requires sensitivity to slight angular variations, which these networks may not adequately model. Lastly, assembly can be viewed as a dynamic, iterative process \cite{zhan2020generative,cheng2023score,li2024category}, where the placement of each part influences the next. In the case of constructing a complex mechanical system, the alignment of base components directly affects how subsequent parts fit together. Yet, current methods typically rely on single-pass predictions, which fail to account for inter-dependencies among parts during sequential assembly, highlighting the need for iterative reasoning or feedback mechanisms to allow for real-time adjustments and refinements.

To address these challenges, we propose a novel approach for network-based absolute pose prediction through the Geometric Point Attention Transformer (GPAT). Our model consists of two key components: the geometric point attention module and the geometric recycling module, both specifically designed to tackle issues related to local geometries, 6-DoF predictions, and dynamic modeling. Unlike traditional attention modules that map each feature into keys, values, and queries to assess their relative importance in the latent space \cite{vaswani2017attention}, our approach enhances the attention score calculation by incorporating geometric pairwise distances and orientations between parts \cite{qin2022geometric}. This enables the model to capture spatial relationships critical for accurate assembly. Furthermore, we directly integrate the rotation and translation vectors of each part into the attention score computation \cite{jumper2021highly}. By considering the local rigid transformations of different parts, we ensure that the placement of each part interacts with its geometric context, resulting in more precise predictions. As we stack attention modules, the rotation and translation vectors are dynamically updated based on learned features, which is particularly advantageous for directly predicting poses without losing 6-DoF geometric characteristics. Additionally, the geometric recycling module allows multiple rounds of prediction, where results from previous steps inform the next round, enabling refinement and correction of earlier predictions. By recycling the predicted 6-DoF features and transforming the point clouds of each part according to the predicted transformations, our network evaluates how well each part is positioned at each iteration and makes further adjustments. Experimental results demonstrate the effectiveness of our model compared to existing methods, showcasing improvements in both the semantic assembly tasks on the PartNet dataset and the geometric assembly tasks on the Breaking Bad dataset. 

To summarize, our main contributions are: 
\begin{itemize} 
\item We design a geometric point attention module that enhances the model's ability to capture local geometric relationships and interactions between parts and poses. 

\item We introduce a geometric recycling scheme that allows for iterative refinement of predictions, significantly improving pose accuracy. 

\item Our overall framework demonstrates strong performance, making it a preferred backbone for future research in shape assembly and other 6-DoF prediction tasks. 
\end{itemize}

\begin{figure*}[t]
    \centering
    \includegraphics[width=\linewidth]{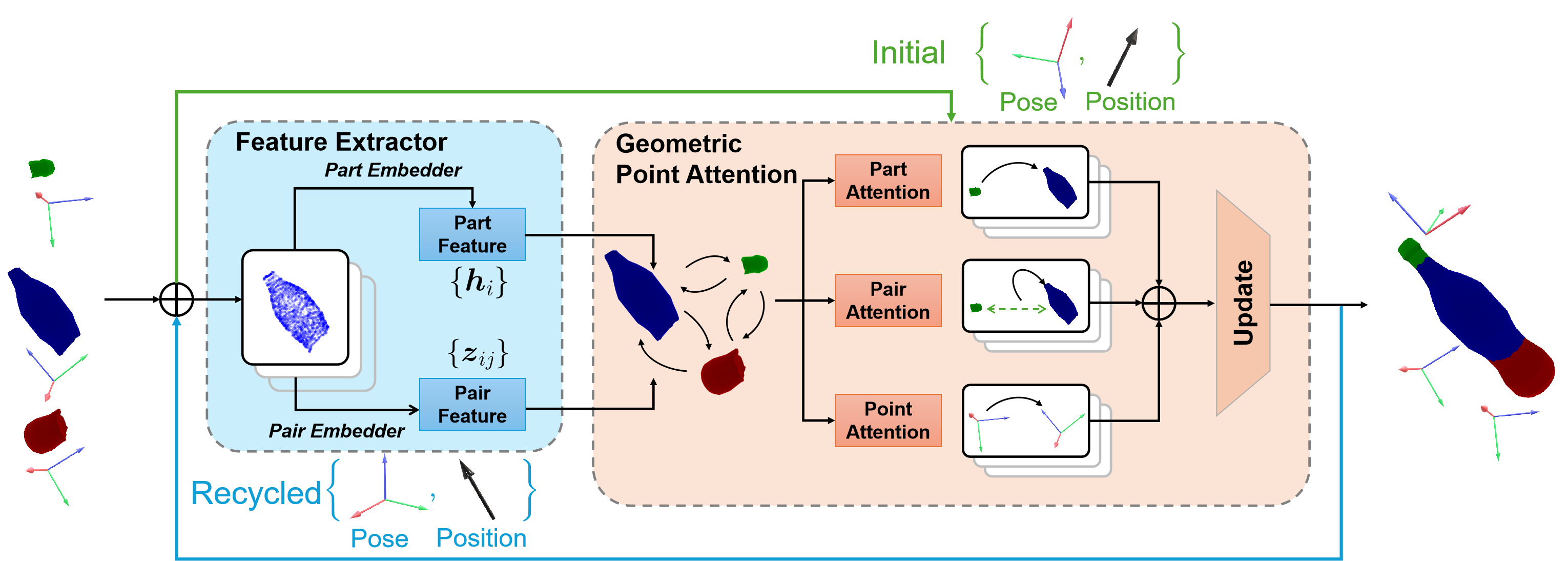} 
    \caption{Overview of our model architecture. Given the point clouds of each part, we first use a feature extractor to generate part features and pairwise features. These features, along with the initial poses, are updated in a stack of geometric point attention modules. The predicted poses and positions are recycled for the next round of predictions in the geometric recycling module.}
    \label{fig:model}
\end{figure*}

\section{Related Work}

\subsection{3D Shape Assembly}

Research in 3D shape assembly focuses on reconstructing complete objects from either predefined semantic parts (part assembly) or fractured pieces (geometric assembly). In the context of part assembly, the large-scale PartNet dataset \cite{mo2019partnet} has facilitated significant progress. For example, Li et al. \cite{li2020learning} assemble parts into a target shape by predicting the translations and rotations of given point clouds using image priors. RGL-Net \cite{narayan2022rgl} and DGL \cite{zhan2020generative} leverage dynamic graph learning and iterative message-passing techniques to merge parts into cohesive structures. However, these methods heavily rely on the availability of semantic segmentation to guide the assembly process. This reliance poses challenges when such labels are missing, especially in geometric assembly tasks, where models must reassemble fractured parts of objects like vases or artifacts based purely on geometric clues. 

For geometric assembly, the Breaking Bad dataset \cite{sellan2022breaking} provides non-semantic fragments, creating a more challenging scenario. In this context, NSM \cite{chen2022neural} prioritizes shape geometries over semantic cues, while Wu et al. \cite{wu2023leveraging} extract geometric features using SE(3)-Equivariant representations. Other approaches explore generative models for fragment reassembly \cite{cheng2023score,scarpellini2024diffassemble}. Additionally, methods have begun to utilize fracture surface features or complete shape templates to streamline assembly, aiming for more general-purpose models \cite{lu2024jigsaw,lu2024jigsaw++}. Techniques such as mapping relative transformations using correspondence alignment estimation have also been proposed \cite{lee20243d}. In contrast, our work focuses on developing a novel network architecture capable of handling both semantic and geometric assemblies. We frame the task as an absolute pose prediction problem, integrating both geometric and contextual information to achieve precise assembly, even in the absence of semantic labels.

\subsection{Transformers}

Transformers, primarily based on self-attention mechanisms, have achieved widespread success in language and vision tasks \cite{vaswani2017attention, achiam2023gpt, dosovitskiy2020image}. In the domain of 3D point cloud processing, early methods mainly leveraged global context by directly utilizing high-level features from point clouds \cite{wang2019deep, yu2021cofinet, huang2021predator}. However, capturing geometric information is equally crucial. This can be achieved through positional embeddings \cite{yang2019modeling, zhao2021point} or by introducing invariant geometric features \cite{qin2022geometric, yu2023rotation}. Nevertheless, the local frames and poses of individual parts are essential for effective part interactions during shape assembly. While recent studies have explored various forms of geometric encoding, directly modeling local poses within the attention module remains a promising yet underexplored direction \cite{jumper2021highly}.

\section{Method}
\label{sec:method}

% fig 1 here

Given an object segmented into $N$ parts, such as the legs and base of a stool, or the shards of a shattered beer bottle, the objective of the shape assembly task is to reassemble these parts to reconstruct the original object with a specified shape. Formally, we represent the parts as $\mathcal{P}=\{\mP_i\}_{i=1}^N$, where each part ${\mP}_i=\{\vx_j \in \mathbb{R}^3\}_{j=1}^{N_i}$ is a point cloud containing $N_i$ points for each part $i$, uniformly sampled from its surface. For tasks involving semantic part assembly (e.g., reconstructing parts of a stool), each part is labeled with a semantic tag indicating its type. In contrast, for purely geometric assembly tasks, only geometric features such as point cloud coordinates are available.

Mathematically, the goal of shape assembly can be formulated as the prediction of the canonical 6-DoF poses $\{\mT_i \in \mathrm{SE}(3)\}_{i=1}^{N}$ for each part, where $\mT_i=(\mR_i \in \mathrm{SO}(3), \vt_i \in \mathbb{R}^3)$ consists of both the rotation matrix and translation vector. Using the predicted pose, each part’s point cloud is transformed to obtain its grounded position: $\mP_i^\text{pred} = \mT_i \circ \mP_i = \mP_i \mR_i + \vt_i$, ultimately reassembling the complete object $\mP^\text{pred} = \{\mT_i\circ\mP_i\}_{i=1}^N$.

Our work focuses on designing a network architecture capable of accurately estimating the absolute poses and capturing the local geometry of each part and pose. To this end, we propose a geometric point attention module, which, together with node and edge attentions, helps better capture the geometric interactions between different parts. Additionally, to enable dynamic reasoning and iterative pose refinement, we introduce a geometric recycling procedure that recursively predicts and refines poses. The overall framework is illustrated in Figure \ref{fig:model}. Initially, features are extracted from the input point clouds and fed into a transformer. Within the transformer, poses and features are iteratively updated by distinct multi-head cross-attention modules, with regularization imposed based on geometric relationships. The output poses serve as predictions and are transformed into recycled features if further refinement is required. In Section \ref{sec:gpa}, we provide details on the Geometric Point Attention Transformer, and in Section~\ref{sec:recycle}, we introduce the novel geometric recycling module.

\begin{figure}[ht]
    \centering
    \includegraphics[width=\linewidth]{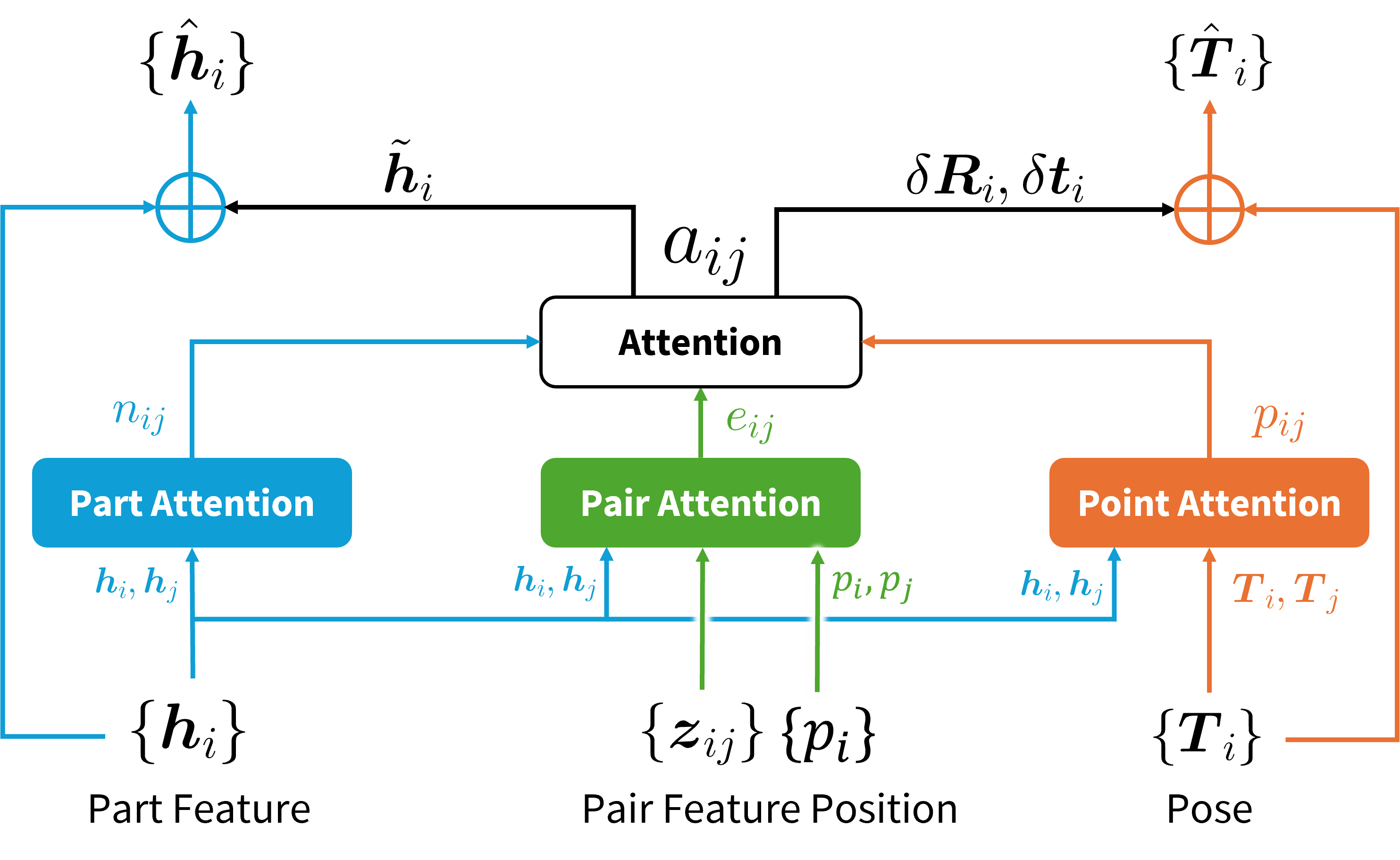}
    \caption{The computation graph of the geometric point attention module. Different input features are fused by the final attention block, with information across parts, Paris, and poses.}
    \label{fig:gpa}
\end{figure}

\subsection{Feature Extraction}
\label{sec:faet}

We begin with a backbone feature extractor, such as PointNet \cite{qi2017pointnet,qi2017pointnet++} or DGCNN \cite{wang2019dynamic}, to capture local hidden geometric features from the point cloud of each part. To provide a shape prior, we also extract a global shape feature by pooling these local geometric features \cite{schor2019componet}. The part-level geometric features $\vh_i^\text{local}$, global feature $\vh^\text{global}$, and recycled geometric features $\vh_i^\text{pos},\vh_i^\text{pose}$ are then concatenated and processed by multi-layer perceptrons (MLPs), referred to as part embedders, to generate the node feature $\vh_i \in \mathbb{R}^d$ for each part. The recycling mechanism is further elaborated in Section \ref{sec:recycle}. In the first round of network prediction, the recycled features are initialized to zero.
\begin{align}
    \vh_i^\text{local}=\mathrm{Backbone}(\mP_i),\quad
    \vh^\text{global}=\frac{1}{N}\sum_{i=1}^{N}\vh_i^\text{local}\\
    \vh_i=\mathrm{MLP}(\mathrm{concat}(\vh_i^\text{local},\vh^\text{global},\vh_i^\text{pos},\vh_i^\text{pose}))
\end{align}
However, relying solely on part-level features for pose prediction is insufficient, as assembly requires capturing meaningful geometric relationships across parts. Therefore, we construct pairwise cross-part features $\vz_{ij} \in \mathbb{R}^d$, which are derived by concatenating the local features $\vh_i^\text{local}$ and $\vh_j^\text{local}$. Along with the recycled geometric features $\vz_{ij}^\text{pos}$, these represent the hidden geometric relations between parts $i$ and $j$. The recycled features are initialized to zero in the first round:

\begin{equation}
    \vz_{ij}=\mathrm{MLP}(\mathrm{concat}(\vh_i^\text{local},\vh_j^\text{local},\vz_{ij}^\text{pos}))
\end{equation}

For the input to the geometric attention module where the 6-DoF pose is explicitly modeled, we initialize the pose of each part to the identity rotation matrix and zero translation vector. This initialization ensures invariance to global rigid transformations, with the pose of each part centered at the origin:  $\mT_i=(\mR=\mI \in \mathrm{SO}(3),\vt = \boldsymbol{0}\in \R^3 )$.
The part features $\vh_i$ and pair features $\vz_{ij}$, along with the initial pose of each part, are then input to the Geometric Attention Transformer.

% start from node and edge feature, copy af2 and geo transformer
\subsection{Geometric Point Attention}\label{sec:gpa}

After extracting features from the point cloud of each part, we use the geometric point attention module to update both the node features and the poses of each part. Current attention modules \cite{vaswani2017attention} can capture global context and cross-part information, which are essential in point cloud modeling. However, these networks tend to ignore the geometric features of part pairs and pose information. Our proposed module not only incorporates high-level node features in \textbf{Part Attention} but also considers high-level pair representations and geometric pair features in \textbf{Pair Attention}. Furthermore, we introduce the invariant \textbf{Point Attention} module to directly model pose information across different parts and update poses across layers in an equivariant way. While our attention module is designed to be multi-layer and multi-head, we omit the layer and head notations for simplicity.

% fig 2 here

\textbf{Part Attention.} The first part of the geometric attention module is part attention, where part features attend to each other to compute relative attention weights, which indicate high-level part interactions and global context extraction. This can also be viewed as a fully connected graph \cite{zhan2020generative,li2024category}. The node features are transformed into query, key, and value vectors, and the squared multiplication between the query and key vectors represents the part-level attention weights:

\begin{align}
    (\vq_i,\vk_i,\vv_i) = (\mW_q\vh_i, \mW_k\vh_i, \mW_v\vh_i),\quad
    n_{ij} = \frac{\vq_i\vk_j^T}{\sqrt{d}}
\end{align}

\textbf{Pair Attention.} After modeling part-level attention, we introduce cross-part pair features, which are incorporated as an additional term in the attention calculation to regulate the cross-part relative weights. First, we transform the input pair feature into the edge attention term as $\vb_{ij} = \mW_b \vz_{ij}$. However, in addition to high-level cross-part representations, the geometric structure between parts should also be included to ensure geometric consistency. This encourages the modeling of dynamic geometric relations between parts, such as the distance and orientation between the feet of a chair, which should stay within an appropriate range when assembled into the complete object \cite{zhan2020generative}. To achieve this, we include geometric invariant distances and orientations between parts in the edge module. Since each part is represented as a point cloud, we first compute the center of mass for each part, which is invariant to global translations:

\begin{equation}
    \vp_i = \frac{1}{N_i}\sum_{j=1}^{N_i}\vx_j
\end{equation}

Next, we compute pairwise distances and triplet-wise dihedral angles between different parts \cite{qin2022geometric,yu2023rotation}. To adopt continuous representations of scalar distances and angles, we use Gaussian radial basis functions \cite{gasteiger2020directional,gasteiger2020fast} to map them into a vector space, given $d$ basis functions:

\begin{align}
    \vd_{ij} &= \mathrm{RBF}(\|\vp_i-\vp_j\|_2) \in \R^d\\
    \vr_{ij} &= \sum_{k=1}^{N}\mathrm{RBF}(\cos{\angle_{ijk} }) \in \R^d
\end{align}

Finally, we combine the high-level edge features with the transformed geometric invariant distance and angle features to obtain the final edge attention term:

\begin{equation}
    \ve_{ij} = \vb_{ij}+\mW_{d}\vd_{ij}+\mW_{r}\vr_{ij}
\end{equation}

\textbf{Point Attention.} For modeling part-level cross-attention, along with cross-part pair attention, we explicitly model the pose information in the attention module, as this is crucial for 6-DoF prediction tasks. Intuitively, the poses between different parts are correlated. For example, when transforming a foot on the bottom of a chair, the transformation of the seat should place it above the foot. Inspired by the success of AlphaFold2 \cite{jumper2021highly}, where each protein residue is associated with a local frame, we leverage an invariant point attention module to model the 6-DoF poses of each part.

We begin by mapping the node features to a set of $N$ virtual 3D vectors in space, which can be interpreted as extracting multiple learnable feature points from each point cloud \cite{yu2021pointr,luo2022equivariant}. For attention calculations, we split these key points into query, key, and value points, with $m \in \{1, \dots, N\}$ denoting different points. Here, $\vec{\vq}_i^m, \vec{\vk}_i^m, \vec{\vv}_i^m$ are 3D vectors in $\mathbb{R}^3$:

\begin{equation}
    (\vec{\vq}_i^m, \vec{\vk}_i^m, \vec{\vv}_i^m) = (\mW_{q}^m\vh_i,\mW_{k}^m\vh_i,\mW_{v}^m\vh_i)
\end{equation}

Since these feature points are embedded in 3D Euclidean space \cite{deng2021vector}, we can directly apply the rigid transformations represented by the pose of each part to them, indicating the relative transformations between different parts and their geometric relations at the local feature level:

\begin{equation}
    p_{ij} = \sum_m\|\mT_i\circ\vq_i^m-\mT_j\circ\vk_j^m\|^2
\end{equation}

By applying pose transformations to the local feature points and computing the L2 norm of the 3D vectors, $p_{ij}$ becomes invariant to global transformations, which is essential for robust and efficient point cloud modeling \cite{deng2021vector,luo2022equivariant,li2022orientation}. Specifically, consider a global transformation $\mT_{\text{global}}$ applied to the entire object consisting of all parts. The poses of each part $\mT_i$ should also be transformed according to this global transformation, while $p_{ij}$ remains the same, as the rigid transformations preserve the L2 norm:

\begin{equation}
    \|\mT_i\circ\vq_i^m-\mT_j\circ\vk_j^m\|^2 = \|\mT_\text{global}\circ(\mT_i\circ\vq_i^m-\mT_j\circ\vk_j^m)\|^2
\end{equation}

\textbf{Feature Update.} After obtaining attention weights from the part, pair, and point attention modules, we use these weights to update part features and poses. The final attention weights are computed by combining the different weights, followed by the softmax function. Here, $n_{ij}$ represents the contributions of global features from neighboring nodes, $e_{ij}$ incorporates edge information and geometric invariant relationships, while $p_{ij}$ encapsulates relative transformation information:

\begin{equation}
    a_{ij} = \mathrm{softmax}(n_{ij} + e_{ij} - p_{ij})
\end{equation}

Note that we add part attention weights and pair attention weights but subtract the point attention weights. This is because $p_{ij}$ essentially measures the distance error between transformed local feature points: a large error suggests that the poses are misaligned, reducing the influence of this term. In contrast, a small error indicates that the part pair is well-aligned, thus contributing more to the feature representation.

The attention weights are used separately to update part, pair, and point features, as they reside in distinct high-level subspaces:

\begin{align}
    \vo_i^{n}&=\sum_{j}a_{ij}\vv_j,\quad
    \vo_i^{e}=\sum_{j}a_{ij}\vz_{ij}\\
    \vec{\vo}_i^{m}&=\mT_i^{-1}\circ\left(\sum_{j}a_{ij}(\mT_i\circ\vec{\vv_j}^{m})\right)
\end{align}

The point feature update remains invariant because the global transformation cancels out, and since the edge attention is conditioned solely on invariant geometric features, the entire update process is invariant.

% Specifically, we have:
% \begin{align}
%     (\mT_\text{global}\circ \mT_i)^{-1}= 
%     % \\ \mT_i^{-1}\circ \mT_\text{global}^{-1}\circ \mT_\text{global}\circ \sum_{j}a_{ij}(\mT_j)\circ \vec{\vv_j}^m= \\ 
%     \mT_i^{-1}\circ \sum_j a_{ij}(\mT_j\circ \vec{\vv_j}^m) 
% \end{align}

We then update the part features by concatenating these updated features, followed by multi-layer perceptrons (MLPs) and residual connections:

\begin{align}
    \tilde{\vh}_i &= \mathrm{MLP}(\mathrm{concat}(\vo_i^{n},\vo_i^{e},\|\vec{\vo}_i^{m}\|_2))\\
    \hat{\vh_i} &= \vh_i +  \tilde{\vh}_i
    \label{eq:update_h}
\end{align}

\textbf{Pose Update.} After updating the node feature using the combined attention weights, we proceed to update the input pose of each part in this attention layer. Unlike other methods that directly predict pose transformations, we estimate the relative transformation from the input pose to the updated pose. This approach not only ensures equivariance but also facilitates dynamic pose updates across attention layers, enabling gradual and easier-to-optimize pose adjustments.

Based on the part feature, we predict the relative rotation and translation. Instead of directly predicting the axis-angle vector or rotation matrix, we predict the unnormalized quaternion representation of the rotation transformation:

\begin{align}
    [b_i,c_i,d_i] = \mathrm{MLP}(\vh_i),\quad \delta\vt_i = \mathrm{MLP}(\vh_i) \\
    \delta\mR_i = \mathrm{quat2rot}\left(\frac{1}{\sqrt{1+b_i^2+c_i^2+d_i^2}},b_i,c_i,d_i\right) \label{eqn:update_r}
\end{align}

We then apply the predicted relative transformation $\delta \mT_i = (\delta \mR_i, \delta \vt_i)$ to the input pose of the attention layer, $\mT_i = (\mR_i, \vt_i)$, to obtain the updated pose $\hat{\mT} = (\hat{\mR_i}, \hat{\vt_i})$, where left-multiplication serves as the standard operation:

\begin{align}
    \hat{\mR_i}= \delta \mR_i \mR_i,\quad 
    \hat{\vt_i}&= \delta \mR_i \vt_i + \delta \vt_i
\end{align}

The pose update is equivariant to global rigid transformations. When a global transformation is applied to the input pose, the predicted relative transformation remains invariant, so the output pose is transformed accordingly, ensuring equivariance.

\subsection{Geometric Recycling} \label{sec:recycle}

To enhance the dynamic geometric reasoning capabilities within the stacked geometric point attention layers, we introduce a technique called Geometric Recycling. This method allows the network to iteratively refine its predictions by incorporating prior results. In complex assembly tasks, using previous predictions as additional input can help the model correct and adjust its prior estimates. Specifically, the outputs of the stacked attention modules—namely, the predicted poses $\mT_i$ and node features $\vh_i$—are fed back into the network as contextual information for further refinement.

\textbf{Position Recycling.} To improve the network's awareness of part positions after each transformation, we apply each part's predicted pose to its point cloud and then re-extract features from this transformed cloud. This process enables the model to better assess how closely the transformed part aligns with the complete structure. First, we apply the predicted pose $\mT_i$ to the input point cloud, and then the backbone network extracts new features for the part. These transformed features are incorporated into the node features:
\begin{align}
    \tilde{\mP}_i = \mathrm{stopgrad}(\mT_i) \circ \mP_i,\quad
    \vh_i^\text{pos} =  \mathrm{Backbone}(\tilde{\mP_i})
\end{align}

Geometric relationships between parts are also critical for assembly. To capture these relationships, we compute the center of mass for each transformed part and calculate the pairwise distances between their centers. These distances are processed using Gaussian radial basis functions (RBFs) to obtain a continuous vector representation, which is then mapped to the edge feature: 

\begin{align}
    \vp_i' = \frac{1}{N_i}\sum_{j=1}^{N_i}\vx_j',\quad
    \vz_{ij}^\text{pos} = \mathrm{RBF}(\|\vp_i'-\vp_j'\|_2) \in \R^d
\end{align}

\textbf{Pose Recycling.} We also recycle the predicted pose itself. The predicted rotation matrix is converted to axis-angle form, and we apply trigonometric functions to the angle vector before concatenating it with the translation vector:

\begin{align}
\vr_i &= \mathrm{mat2axis}(\mathrm{stopgrad}(\mR_i))\\
\vh_i^\text{pose} &= \mathrm{MLP}(\mathrm{concat}(\sin(\vr_i),\cos(\vr_i),\mathrm{stopgrad}(\vt_i)))
\label{eq:recycle_pose}
\end{align}

% axis angle + mlp

\textbf{Training and Inference.}  Recycling enables the network to iteratively process updated versions of the input features without significantly increasing training time or model size. Given a recycling number $N_{\text{recycle}}$ (the number of recursive rounds), we adjust the training and inference procedures accordingly.

During training, we optimize the network to be robust across multiple iterations. To simplify training, we employ stop-gradient operations, which prevent gradient backpropagation across rounds; thus, recycled features provide additional information without influencing optimization directly. Loss is computed only on the output of the final round, effectively optimizing the average loss across different recycling numbers.

During inference, recycling forms a recurrent network with shared weights that iteratively refines the output. We use a fixed recycling count during inference, which may exceed the range used during training to allow for further refinement.

% training, sample, expectation

% inference, fix inference step, 

% \subsection{Loss Functions}
\section{Experiment}

\begin{table*}[t]
\centering
\label{tab:semantic}
\resizebox{0.85\linewidth}{!}{%
\begin{tabular}{l|ccc|ccc|ccc}
\toprule
 & \multicolumn{3}{c|}{Shape Chamfer Distance ↓} & \multicolumn{3}{c|}{Part Accuracy ↑} & \multicolumn{3}{c}{Connectivity Accuracy ↑} \\
Method & Chair & Table & Lamp & Chair & Table & Lamp & Chair & Table & Lamp \\
\midrule
B-Global & 0.0146 & 0.0112 & 0.0079 & 15.7 & 15.37 & 22.61 & 9.90 & 33.84 & 18.6 \\
B-LSTM & 0.0131 & 0.0125 & 0.0077 & 21.77 & 28.64 & 20.78 & 6.80 & 22.56 & 14.05 \\
B-Complement & 0.0241 & 0.0298 & 0.0150 & 8.78 & 2.32 & 12.67 & 9.19 & 15.57 & 26.56 \\
DGL & 0.0091 & 0.0050 & 0.0093 & 39.00 & 49.51 & 33.33 & 23.87 & 39.96 & 41.70 \\
\midrule
GPAT & 0.0082 & 0.0043 & 0.0099 & 43.29 & 51.64 & 34.33 & 29.23 & 41.04 & 48.10 \\
GPAT w/o Attention & 0.0098 & 0.0056 & 0.0112 & 36.70 & 45.32 & 28.90 & 23.10 & 37.58 & 40.25 \\
GPAT w/o Recycle & 0.0087 & 0.0049 & 0.0105 & 41.10 & 49.20 & 32.00 & 27.50 & 39.20 & 44.70 \\
\bottomrule
\end{tabular}%
}
\caption{Quantitative comparisons and ablation study on PartNet dataset for semantic Assembly.}
\end{table*}

\begin{figure*}[t] % 使用 figure* 使图片横跨双栏
    \centering
    \includegraphics[width=0.85\linewidth]{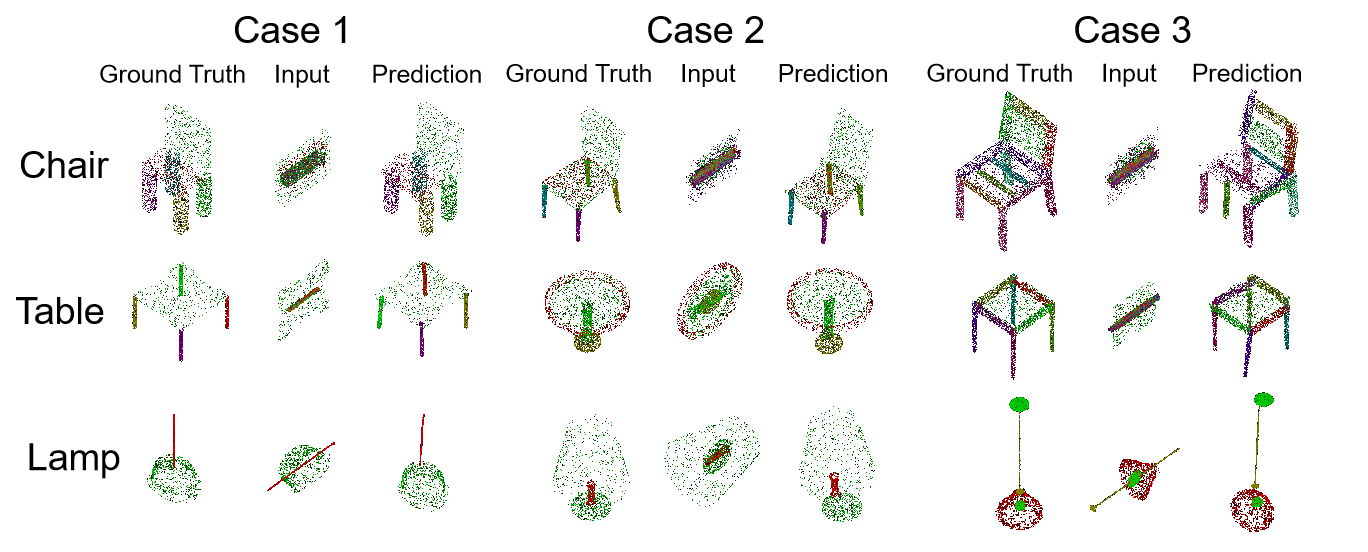} % 调整图片大小
    \caption{Qualitative results of part assembly using predicted poses from GPAT for semantic assembly.}
    \label{fig:semantic}
\end{figure*}

\subsection{Semantic Assembly}

\textbf{Dataset.} Following \cite{zhan2020generative}, we use the three largest furniture categories in PartNet \cite{mo2019partnet}: chairs, tables, and lamps, which consist of fine-grained shapes with part segmentation labels (e.g., the legs of a table or the backrest of a chair). These categories contain a total of $6,323$, $8,218$, and $2,207$ shapes, respectively. Each shape is composed of multiple segmented parts, and each part is represented by $1,000$ points sampled from its mesh.

\textbf{Metrics.} We use three distance metrics to evaluate assembly quality across different categories \cite{zhan2020generative, cheng2023score, li2024category}. The shape chamfer distance quantifies overall assembly quality by calculating the chamfer distance between the ground truth shape and the predicted assembly, which is composed of transformed parts. Part accuracy measures how well each part matches its ground truth counterpart, indicating the percentage of parts that fall within a specified distance threshold. Additionally, connectivity accuracy evaluates how well parts are connected in the assembled shape by checking pairs of contact points. We set the minimum chamfer distance threshold to $0.01$.

\textbf{Baselines.} In line with prior work \cite{zhan2020generative}, we compare our method with several baselines trained using the same losses and hyperparameters. Similar to CompoNet \cite{schor2019componet} and PAGENet \cite{li2020learning}, \textbf{B-Global} enhances part features with global context and directly decodes poses using MLPs. \textbf{B-LSTM} employs a bidirectional LSTM backbone, as used in PQNet \cite{wu2020pq}, to sequentially predict part poses based on prior estimations. \textbf{B-Complement} retrieves candidate parts from a database to assemble the complete shape, sequentially predicting poses for each part \cite{sung2017complementme}. Finally, \textbf{DGL} leverages dynamic graph learning with an iterative graph neural network to refine shape assembly based on a set of parts \cite{zhan2020generative}.

\textbf{Results.} The quantitative results of our method and the baselines are summarized in \cref{tab:semantic}. Our approach consistently outperforms the baselines across nearly all metrics, particularly in terms of part accuracy and connectivity accuracy. This demonstrates that our model can produce high-quality assemblies at the part level and achieve precise alignment in the overall shape. As shown in \cref{fig:semantic}, the assembled shapes generated by our method closely match the ground truth objects in terms of overall structure. However, we observe that some parts exhibit slight deviations in their rotational axes, suggesting challenges in precisely estimating rotations. These rotational discrepancies are more pronounced in parts with complex orientations, highlighting a potential area for improvement in refining axis alignment during the assembly process. Nonetheless, the strong performance of our model confirms its effectiveness in both part-level precision and overall shape reconstruction.

\subsection{Geometric  Assembly}

\begin{table*}[t]
    \centering
    \resizebox{0.8\linewidth}{!}{%
    \begin{tabular}{l|ccc|cc|cc}
        \toprule
        \multirow{3}{*}{Method} 
        & \multicolumn{3}{c|}{Rotation} 
        & \multicolumn{2}{c|}{Translation} 
        & \multicolumn{2}{c}{Accuracy} \\
        \cmidrule(lr){2-4} \cmidrule(lr){5-6} \cmidrule(lr){7-8}
        & RMSE (R) $\downarrow$ & MAE (R) $\downarrow$ & GD (R) $\downarrow$ 
        & RMSE (T) $\downarrow$ & MAE (T) $\downarrow$ 
        & CD $\downarrow$ & PA $\uparrow$ \\
        & (degree) & (degree) &  
        & ($\times 10^{-2}$) & ($\times 10^{-2}$) 
        & ($\times 10^{-3}$) & (\%) \\
        \midrule
        Global  & 82.6 & 70.6 & 2.17  & 14.8 & 11.8 & 25.9 & 21.2 \\
        LSTM  & 85.5 & 74.7 & 2.21 & 16.2 & 13.7 & 26.2 & 19.2 \\
        DGL  & 81.6 & 69.8 & 2.12 & 15.8 & 12.6 & 25.3 & 23.9 \\
        NSM  & 86.5 & 74.2 & 2.24 & 16.8 & 15.7 & 26.9 & 16.2 \\
        SE(3)-Equi & 78.2 & 65.8 & 2.04 & 14.8 & 24.5 & 14.7 & 24.8 \\
        \midrule
        GPAT & 79.3 & 66.4 & 2.08
        & 14.4 & 11.1 & 23.0 & 30.2 \\
        GPAT w/o Attention & 81.5 & 68.9 & 2.13
        & 14.2 & 11.3 & 23.6 & 26.3 \\
        GPAT w/o Recycle & 80.0 & 67.0 & 2.10
        & 14.6 & 11.8 & 23.4 & 29.5 \\
        \bottomrule
    \end{tabular}
    }
    \caption{Quantitative comparisons and ablation study on the everyday object subset of the Breaking Bad dataset for geometric assembly.}
    \label{tab:geometric}
\end{table*}

\textbf{Dataset.} We use the Breaking Bad dataset \cite{sellan2022breaking}, which contains objects that are irregularly broken into multiple fragmented pieces through synthetically generated, physically plausible decompositions. This results in more complex geometries and a higher number of parts per object. For our experiments, we use the ``everyday'' subset, which includes $20$ categories, comprising $34,075$ fracture patterns from $407$ objects for training and $7,679$ fracture patterns from $91$ objects for testing.

\textbf{Metrics.} Similar to the semantic assembly task, we use chamfer distance (\textbf{CD}) between the ground truth shape and the predicted assembly to evaluate overall reconstruction quality. Part accuracy (\textbf{PA}) is used to assess assembly quality at the individual part level. Given the critical importance of accurate pose estimation for assembling fractured fragments \cite{wu2023leveraging}, we also evaluate prediction errors for both rotation and translation. Specifically, we compute the root mean squared error (\textbf{RMSE}) and mean absolute error (\textbf{MAE}) between the predicted poses and ground truth values. Additionally, we include geodesic distance (\textbf{GD}) to measure rotational alignment accuracy.

\textbf{Baselines.}  As in the semantic assembly task, we include \textbf{Global}, \textbf{LSTM}, and the iterative \textbf{DGL} GNN-based method as our baselines. Additionally, we evaluate two more advanced approaches: Neural Shape Mating \cite{chen2022neural}, which uses a transformer-based feature extractor and implicit shape reconstruction for pairwise pose prediction, and SE(3)-Equi \cite{wu2023leveraging}, which leverages both equivariant and invariant features through a vector-based network \cite{deng2021vector}.

\begin{figure}[h] % 使用 figure* 使图片横跨双栏
    \centering
    \includegraphics[width=\linewidth]{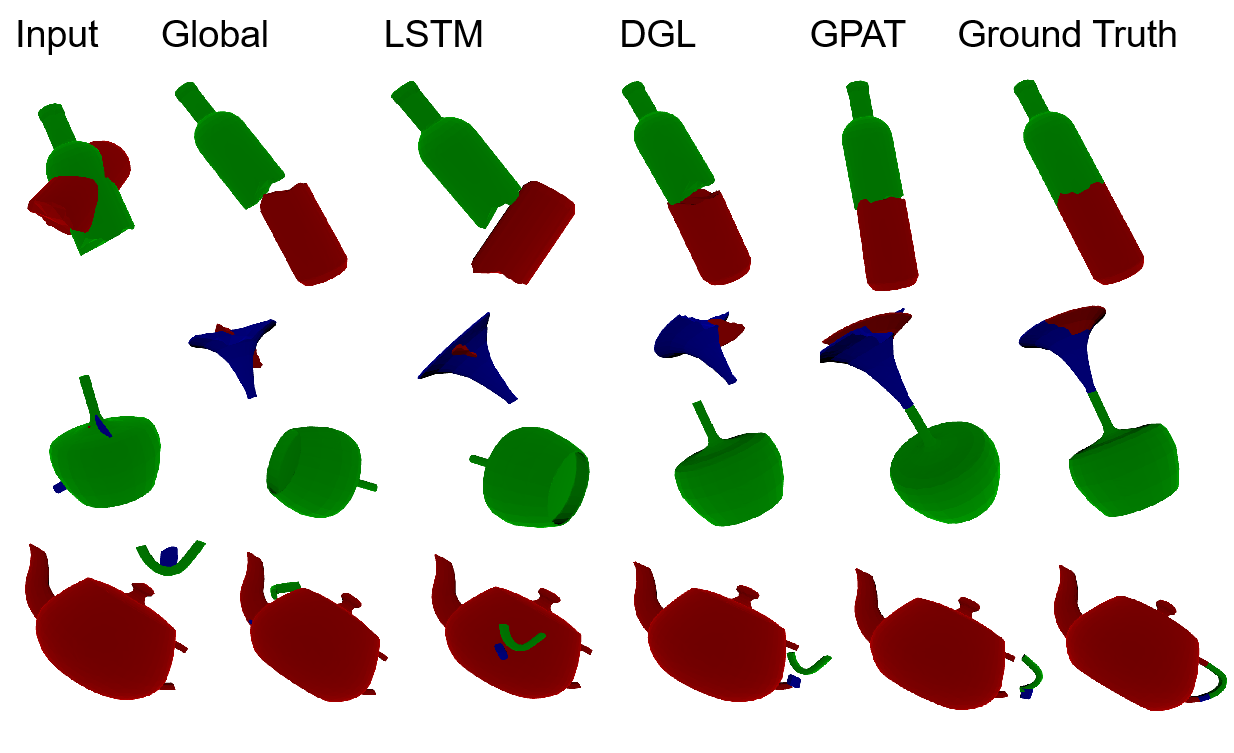} % 调整图片大小
    \caption{Qualitative comparison between GPAT and other baselines for geometric assembly. GPAT outperforms across different object shapes and numbers of fragments.}
    \label{fig:geometric}
    % \vspace{-2em}
\end{figure}

\textbf{Results.} The quantitative results for the geometric assembly task on the Breaking Bad dataset are presented in \cref{tab:geometric}. Our method outperforms all baselines in translation pose prediction but is slightly less accurate than the vector-based network in rotation estimation. Nevertheless, our approach achieves comparable or superior results in part assembly and overall shape reconstruction, indicating that it better captures the consistency and interplay between predicted rotation and translation poses, thanks to the invariant point attention module. Moreover, incorporating both global and local geometric relationships enables our model to effectively balance these two critical components, leading to coherent part alignments. As shown in \cref{fig:geometric}, our method consistently produces shapes that closely resemble the ground truth across various object categories and different numbers of fractured parts. However, we observe some limitations in fine-grained detail modeling by absolute pose prediction networks. Specifically, in the assembled results, some parts are positioned too close, leading to mesh clashes, while others are spaced too far apart, causing mismatches. This suggests that further improvements could be made by incorporating post-processing techniques, such as point matching or texture mapping \cite{lu2024jigsaw,lee20243d}, to refine part alignments and enhance overall assembly accuracy.

\subsection{Ablation Studies} To assess how different components of our model contribute to its superior performance, we perform ablation studies on the geometric point attention module and geometric recycling module, denoted as \textbf{GPAT w/o Attention} and \textbf{GPAT w/o Recycle}. In \textbf{GPAT w/o Attention}, we retain only node attention while removing both edge and point attention. In \textbf{GPAT w/o Recycle}, the model predicts poses in a single pass, with recycling features set to zero. As shown in \cref{tab:semantic} and \cref{tab:geometric}, both the attention module and recycling procedure are critical for accurate pose estimation. The geometric point attention module plays a key role in improving prediction accuracy by explicitly capturing local transformations and spatial relationships between parts. This allows the model to better understand complex geometric dependencies and interactions, leading to more accurate part placements. The recycling procedure further enhances performance by iteratively refining predictions. By feeding back the predicted poses into the network, the model can correct errors and adjust the assembly in subsequent rounds, improving overall alignment and robustness. Together, these components ensure that our model excels in both part-level assembly accuracy and global shape reconstruction.
\section{Conclusion}
In this paper, we introduced the Geometric Point Attention Transformer  to tackle the challenging problem of 3D shape assembly, focusing on part-level 6-DoF pose prediction. By integrating global shape information, local geometric features, and pose transformations directly into the attention mechanism, GPAT effectively models spatial relationships. The addition of a geometric recycling scheme allows for iterative refinement, leading to improved assembly accuracy. Our experiments on PartNet and Breaking Bad datasets demonstrate that GPAT outperforms existing methods, highlighting its potential as a robust solution for both semantic and geometric assembly tasks and a foundation for future research in complex 3D reconstruction.

\newpage

{
    \small
    \bibliographystyle{ieeenat_fullname}
    \bibliography{main}
}

\clearpage
\setcounter{page}{1}
\maketitlesupplementary

\section{Implementation Details}
We provide detailed information about the implementation of our geometric point attention transformer and the baseline models, including the network architecture (\cref{sec:network}), baseline specifications (\cref{sec:baseline}), loss functions (\cref{sec:loss}), and evaluation metrics (\cref{sec:eval}).

\subsection{Network Architecture}
\label{sec:network}

The overview of our model architectures is shown in \cref{fig:model}, here we demonstrate more details of each part in our model.

In the feature extractor \cref{sec:faet}, we use PointNet as the backbone network to extract hidden features from each point cloud of each part. Each point cloud contains 1,000 points and maps the 3D coordinates $\vp_j \in \R^3$ to a hidden vector $\vh_i^\text{local} \in \R^{d}$. Here, we set the hidden dimension to $d=128$. The global feature $\vh^\text{global} \in \R^{d}$ is obtained by averaging the hidden vectors from all parts. Next, the Part Embedder and Pair Embedder, both implemented as 3-layer Multi-Layer Perceptrons (MLPs) with ReLU activations, separately transform the hidden features into Part Features and Pair Features.

In the geometric attention module \cref{sec:gpa}, we use 12 attention heads, with each head having hidden dimensions $d=32$ for the query, key, and value vectors. For point attention, we use $N=12$ virtual points. The updated features from each attention head and each virtual point are concatenated and transformed into the updated part feature (\cref{eq:update_h}). This updated feature is then used to predict relative rotation and translation (\cref{eqn:update_r}). We stack $4$ layers of attention modules.

In the geometric recycling module \cref{sec:recycle}, we set $N_\text{recycle}=4$ and use radial basis function (RBF) kernels and Euler angle representations to embed distances and rotation matrices. The hidden dimension is set to $d=128$, and the MLP used in \cref{eq:recycle_pose} is a 3-layer network with ReLU activations.

\paragraph{RBF function} %随便写一下
Instead of directly feeding the distance and angles as input features like \cite{deng2018ppfnet}, we follow \cite{schutt2017schnet,gasteiger2020directional} to apply deterministic embedding functions to extract more comprehensive information for the distance and angles. Specifically, radial basis functions (RBFs) are used to embed the distance:
\begin{equation}
    \mathrm{RBF}(d)_n=\sqrt{\frac{2}{c}}\frac{\sin(\frac{n\pi}{c}d)}{d}
\end{equation}
where $c$ is the cut-off distance. Similarly, the RBFs for angles are defined based on the 2D spherical Fourier-Bessel basis, as introduced in \cite{gasteiger2020directional}:
\begin{equation}
    \mathrm{RBF}(\alpha)_{\ell}=Y^0_\ell(\alpha)
\end{equation}
where $Y^0_\ell$ is the spherical harmonics of order 0. Such RBFs can better capture the frequency information in the distances and angles, leading to more comprehensive geometric representations.

\paragraph{mat2axis function} The $\mathrm{mat2axis}$ function in Eq. (\cref{eq:recycle_pose}) transforms the rotation matrix $\mR \in \mathbb{R}^{3 \times 3}$ to its Euler angle representation $\vr = (\phi, \theta, \psi)$, where:
\begin{align}
    \theta &= \arctan2\left(-R_{31}, \sqrt{R_{11}^2 + R_{21}^2}\right), \\
    \phi &= \arctan2\left(R_{32}, R_{33}\right), \\
    \psi &= \arctan2\left(R_{21}, R_{11}\right).
\end{align}

\paragraph{quat2rot function} % af2
The $\mathrm{quat2rot}$ function in Eq. (\cref{eqn:update_r}) converts the quaternion representation $q=a+b\ri+c\rj+d\rk\in\mathbb{H}$ of a rotation into the corresponding rotation matrix $\mR \in \mathbb{R}^{3 \times 3}$:

\begin{align}
    \omega&=2\arctan2(\sqrt{b^2+c^2+d^2},a),\\
    \vr&=\frac{\omega}{\sqrt{b^2+c^2+d^2}}(b,c,d):=(\phi, \theta, \psi),\\
    \mK&=\begin{bmatrix}
0 & -\psi & \theta \\
\psi & 0 & -\phi \\
-\theta & \phi & 0 
\end{bmatrix},\\
    \mR&=\mI+(\sin \omega) \mK +(1-\cos\omega)\mK^2.
\end{align}
where in the last equation we utilize the Rodrigues' rotation formula to reconstruct the rotation matrix from the corresponding rotation vector.

\subsection{Baseline Implementation}
\label{sec:baseline}
\paragraph{Global}
Following \cite{zhan2020generative,sellan2022breaking}, we first extract the part feature for each input point cloud and the global feature using the GNN feature extractor. Then, we concatenate the global feature with each part feature and apply a shared weight MLP network to regress the SE(3) pose for each input point cloud. Such an approach is similar to CompoNet \cite{schor2019componet} and PAGENet \cite{li2020learning}.

\paragraph{LSTM}
Following \cite{zhan2020generative,sellan2022breaking}, instead of leveraging a graph structure to encode and decode part information jointly, a bidirectional LSTM \cite{schuster1997bidirectional} module similar to PQ-Net \cite{wu2020pq} to sequentially estimate the part pose. This resembles the process of sequential decision-making when humans perform shape assembly.

\paragraph{Complement}
ComplementMe \cite{sung2017complementme} studies the task of synthesizing 3D shapes from a big repository of parts and mostly focuses on retrieving part candidates from the part database. Following \cite{zhan2020generative}, we modify the setting to our case by limiting the part repository to the input part set and sequentially predicting a part pose for each part.

\paragraph{DGL}
Dynamic graph learning (DGL) \cite{zhan2020generative} applies an iterative graph neural network to refine shape assembly based on a set of parts, where GNNs encode part features via edge relation reasoning and node aggregation modules. On the Breaking Bad dataset, following \cite{sellan2022breaking}, we remove the node aggregation operation designed for handling geometrically equivalent parts in DGL since every piece in this dataset has a unique shape geometry.

\paragraph{NSM}

Neural Shape Mating \cite{chen2022neural} couples the training of pose estimation with an implicit shape reconstruction task, using signed distance functions (SDFs) and a discriminator for learning shape priors, and applies a transformer regressor for estimating poses of each part.

\paragraph{SE(3)-Equi}

\citet{wu2023leveraging} studies the equivariant issue in 3D geometric assembly, they use vector neuron networks to extract geometric features from each point cloud, and proposes canonical part reconstruction loss and adversarial training scheme to make the network more robust to global transformations.

\subsection{Loss Functions}
\label{sec:loss}
% see in breaking bad, appendix D
We use a similar training loss function following the original Breaking Bad paper \cite{sellan2022breaking}. The poss regression loss $\mathcal{L}_\text{pose}$ for the  output prediction $(\hat{\mR},\hat{\vt})$ is defined as
\begin{equation}
    \mathcal{L}_\text{pose}=\sum_{i=1}^N\|\vt_i-\hat{\vt}_i\|_2^2 +\lambda_\text{rot}\|\mR_i^\top\hat{\mR}_i-\mI\|_2^2
\end{equation}
% As the stop-gradient operation is performed on each round of recycling, we average all the recycling losses together as the final pose regression loss as
% \begin{equation}
%     \mathcal{L}_\text{pose}=\frac{1}{N_\text{recycle}}\sum_{k=1}^{N_\text{recycle}}\mathcal{L}^k_\text{pose}.
% \end{equation}
The Chamfer distance loss $\mathcal{L}_\text{chamfer}$ measures the chamfer distance between the predicted pose-transformed point clouds and the ground truths, as well as the predicted assembly and the ground truth. It is defined as
\begin{equation}
    \mathcal{L}_{\text {chamfer }}=\sum_{i=1}^{N} \mathrm{CD}\left(\mR_{i} \hat{\mP}_i, \hat{\mR}_{i} \hat{\mP}_{i}\right)+\lambda_{\text {shape }} \mathrm{CD}(\mP, \hat{\mP}) .
\end{equation}
The point-to-point MSE loss $\mathcal{L}_{\text {point }}$ measures the point-wise errors and has been demonstrated to help improve rotation prediction. It is defined as the L2 distance between point clouds transformed by the predicted rotation and by the ground-truth rotation, respectively, as
\begin{equation}
    \mathcal{L}_{\text {point }}=\sum_{i=1}^{N} \sum_{j}\|\mR_{i} \hat{\mP}_{i}^{j}-\hat{\mR}_{i} \hat{\mP}_{i}^{j}\|_{2}^{2}.
\end{equation}

In this way, the total loss $\mathcal{L}$ is trained by optimizing the following objective function with
hyperparameters $\lambda_\text{chamfer}, \lambda_\text{point}$ balancing the three loss terms and aggregated across all recycling iterations:
\begin{equation}
    \mathcal{L} = \mathcal{L}_\text{pose}+\mathcal{L}_\text{chamfer}+\mathcal{L}_\text{point}.
\end{equation}

\subsection{Evaluation Metrics}
\label{sec:eval}

% % part net: Generative 3D Part Assembly via
%  Dynamic Graph Learning, sec 4.3

% % breaking bad: 3D Geometric Shape Assembly via Efficient Point Cloud Matching, appendix D

For semantic assembly, We follow \cite{sellan2022breaking,zhan2020generative} to use the shape chamfer distance (\textbf{CD}) and part accuracy (\textbf{PA}) as our evaluation major metrics. The chamfer distance $\mathrm{CD}(P, Q)$ between two point clouds $P$ and $Q$ is defined as:
\begin{equation}
    \mathrm{CD}(P, Q)=\sum_{x \in P} \min _{y \in Q}\|x-y\|_{2}^{2}+\sum_{y \in Q} \min _{x \in P}\|y-x\|_{2}^{2} .
\end{equation}
The shape chamfer distance $\mathrm{CD}(\mP,\hat{\mP})$ is computed between the predicted assembly $\hat{\mP}$ and the ground-truth assembly $\mP$.

Part accuracy measures the percentage of parts whose chamfer distance to ground truth is less than a threshold and is defined as
\begin{equation}
    \mathrm{PA}=\frac{1}{N} \sum_{i=1}^{N} \mathbbm{1}\left(\mathrm{CD}(\mP_{i}, \hat{\mP}_{i})<\tau\right)
\end{equation}
where we set $\tau=0.01$ following \cite{sellan2022breaking}. 

In \cite{zhan2020generative}, the connectivity accuracy (\textbf{CA}) is proposed to further evaluate how well the parts are connected in the assembled shape in addition to PA which considers each part separately. In short, for each connected pair of parts, the coordinates of the closest contact points $\vp_{ij},\vp_{ji}$ in the part frame $i,j$ are transformed to their corresponding local canonical part space as $\vc_{ij},\vc_{ji}$. CA is then defined as
\begin{equation}
    \mathrm{CA}=\frac{1}{N_\text{pair}} \sum_{i,j} \mathbbm{1}\left(\|\vc_{ij}-\vc_{ji}\|_2^2<\tau\right)
\end{equation}
where we set $\tau=0.01$ following \cite{zhan2020generative}. In other words, CA evaluates the percentage of correctly connected parts.

For geometric assembly, additionally, we follow \cite{sellan2022breaking} to calculate the root mean square error (\textbf{RMSE}) and the mean absolute error (\textbf{MAE}) between the predicted rotation $\hat{\mR}$ and translation $\hat{\vt}$,and ground-truth rotation $\mR^{gt}$ and translation $\vt^{gt}$, where the rotation is represented using Euler angles $(\phi, \theta, \psi)$. 
\begin{align}
    \mathrm{MAE}(\mR)&=\frac{1}{3}\|\hat{\mR}-\mR^{gt}\|_1 \\
    \mathrm{RMSE}(\mR)&=\frac{1}{\sqrt{3}}\|\hat{\mR}-\mR^{gt}\|_2 \\
    \mathrm{MAE}(\vt)&=\frac{1}{3}\|\hat{\vt}-\vt^{gt}\|_1 \\
    \mathrm{RMSE}(\vt)&=\frac{1}{\sqrt{3}}\|\hat{\vt}-\vt^{gt}\|_2 \\
\end{align}
The average errors of geodesic distance on SO(3) between the ground truth and the predicted rotations (\textbf{GD}) are also provided to offer a more comprehensive evaluation of the rotation quality, as RMSE and MAE do not necessarily hold the property of a metric. It is well-known that the rotation group SO(3) is also a Riemannian manifold equipped with a modified version of the canonical Frobenius inner product
\begin{equation}
    \langle \mA_1,\mA_2\rangle_{\mathrm{SO}(3)}=\frac{1}{2}\langle\mA_1,\mA_2\rangle_F=\frac{1}{2}\Tr(\mA_1^\top \mA_2)
\end{equation}
where $\mA_1,\mA_2\in\mathfrak{so}(3)$ denote elements in the tangent space of skew-symmetric matrices, i.e., the Lie algebra of SO(3).
The geodesic distance between the predicited rotamtion matrix and the ground truth rotation matrix can be calculated as
\begin{equation}
    d(\Tilde{\mR},\mR^{gt})=\|\log (\Tilde{\mR}^\top \mR^{gt})\|_F^2
\end{equation}
where $\log$ is the matrix logarithm that maps $\mR\in \mathrm{SO}(3)$ to its Lie algebra $\mA\in \mathfrak{so}(3)$, and $\|\mA\|_F^2=\langle\mA,\mA\rangle_F$ is the canonical Frobenius norm.

\end{document}